\documentclass[runningheads]{llncs}
\usepackage{multirow}
\usepackage{multicol}
\usepackage{booktabs}
\usepackage[T1]{fontenc}
\usepackage{graphicx}
\usepackage{cite}
\usepackage{url}
\usepackage[nolist]{acronym}
\usepackage{array}
\usepackage{tikz}

\usepackage{xcolor}
\newcommand{\rev}[1]{\textcolor{black}{#1}}
\begin{document}

\title{Towards Real-World Applications \\ with an Autonomous Powered Wheelchair}
\titlerunning{Towards Real-World Applications with an Autonomous Powered Wheelchair}
\author{Simone Arreghini\inst{1} \and
Alessandro Giusti\inst{1} \and
Alex Bordini\inst{2} \and
Enrico Ferrara\inst{2} \and
Giovanni Fulgoni\inst{2} \and
Antonio Paolillo\inst{1}}
\authorrunning{S. Arreghini et al.}
\institute{Dalle Molle Institute for Artificial Intelligence (IDSIA), USI-SUPSI,
\\ Lugano, Switzerland. Email: \email{name.surname@idsia.ch}
\and
Genny Factory, Sant'Antonino, Switzerland. \\ Email: \email{name.surname@gennyfactory.ch}}

\maketitle%
\begin{abstract}
Wheelchair users call for assistive mobility systems that provide active support, adapt to dynamic environments, and are intuitive and user-friendly.
However, powered wheelchairs typically still provide limited autonomy and lack effective integration with advanced perception and navigation capabilities, particularly in complex real-world environments.
%
% In this paper, we present a preliminary study toward an autonomous powered wheelchair for real-world assistive mobility applications.
\rev{This paper presents a preliminary study toward autonomous powered wheelchairs for real-world assistive mobility. We introduce a proof-of-concept prototype that integrates autonomous perception, gesture-based interaction, and navigation on a commercially available self-balancing powered wheelchair.}
The proposed system builds upon Genny Zero, a commercial self-balancing wheelchair that enables hands-free and intuitive operation through body-weight shifting. 
To extend its capabilities toward autonomous operation, we integrate an RGB-D camera for human-aware perception and interaction, together with a LiDAR sensor for localization and navigation.
\rev{We demonstrate the integrated system in two assistive applications}: (i) hailing, allowing users to call the wheelchair from a distance; and (ii) people-following, where the wheelchair follows a person using leader-follower strategies\rev{, including a constrained indoor navigation example.}
%
% These results highlight the potential of combining autonomous robotics with assistive mobility platforms, paving the way for a novel intelligent, accessible, and user-centered mobility solutions.
\rev{The results highlight the potential of combining autonomous robotics with assistive mobility platforms, while also showing the feasibility of the proposed integration and identifying the main technical challenges that must be addressed before moving toward user-ready, accessible, and intelligent mobility solutions.}
A video demonstrating the experimental setup and results is available at: \url{https://youtu.be/LVAix_Qx7bM}.


\end{abstract}

\section{Introduction}

Today, there is a clear and growing demand for assistive mobility solutions that can actively support users~\cite{who:report:2022}. This demand is further driven by aging populations, which require efficient, easy-to-use, and personalized mobility systems~\cite{bonaccorsi2016cloud}.
Significant research efforts have been devoted to improving accessibility and independence for wheelchair users. 
In this context, powered wheelchairs equipped with autonomous robotic capabilities represent a promising and interesting research direction to address these societal needs.

Autonomous mobile robots have reached a high level of technological readiness, with mature applications in industrial logistics and controlled indoor environments~\cite{boysen2019warehousing}. 
In the domain of assistive mobility, instead, various powered wheelchair systems have been proposed\cite{daav2026, omeo2026, scewo2026}; however, these typically involve only limited levels of autonomy.
Overall, existing solutions often fail to effectively integrate advanced autonomous robotics with assistive mobility platforms into a single, user-friendly system. 
This gap becomes even more pronounced in-the-wild scenarios, such as navigation in public and crowded environments, where additional complexities and challenges arise~\cite{Salvini:soro:2022,Mavrogiannis:thri:2023}.

In this paper, we present a preliminary study toward a novel autonomous powered wheelchair designed for real-world applications, and demonstrate the potential of our approach as a future assistive mobility system.
Our system is built upon the Genny Zero platform, a self-balancing powered wheelchair manufactured by Genny Factory, described in Sec.~\ref{sec:system}. 
Genny Zero is a mobility device controlled through intuitive body-weight shift, which enables hand-free operation, greatly improving the users' independence and quality of life.
Despite being a mature and widely appreciated device, the Genny Zero lacks exteroceptive sensing and autonomous capabilities. 
To address this limitation, we propose the integration of an advanced robotic perception pipeline which, paired with intelligent control, enables safe assistive behavior and shared autonomy.
Specifically, we equip the platform with an RGB-D camera to support human-aware perception~\cite{shotton2013real} and human–robot interaction (HRI)~\cite{Abbate:ras:2024}, and a LiDAR sensor for robust mobility and navigation~\cite{Leong:access:2024}.
We evaluate the effectiveness of the proposed system in two application domains, as detailed in Sec.~\ref{sec:applications}. 
The first considers a people-following mode, in which the wheelchair tracks and follows a user using perception-driven navigation strategies. 
Similar concepts have been explored in assistive robotics and shared autonomy systems~\cite{Antonucci:auro:2023,Liao:ral:2025}, particularly in structured environments such as hospitals and airports~\cite{Salimpour:arxiv:2025}.
The second application is hailing, which allows a user to call the wheelchair from a distance, thereby increasing convenience and independence. 
Such functionality is consistent with recent developments in smart wheelchairs that support remote control~\cite{Jittrapirom:up:2017}.  

Section~\ref{sec:discussion} discusses the current state of the proposed system and its limitations, along with our planned future work toward a fully autonomous powered wheelchair. 
Finally, Sec.~\ref{sec:conclusion} concludes the paper with final remarks.
\section{System and experimental setup}\label{sec:system}
\subsection{Genny Zero platform}

\begin{figure}
    \centering
    \includegraphics[width=0.495\linewidth]{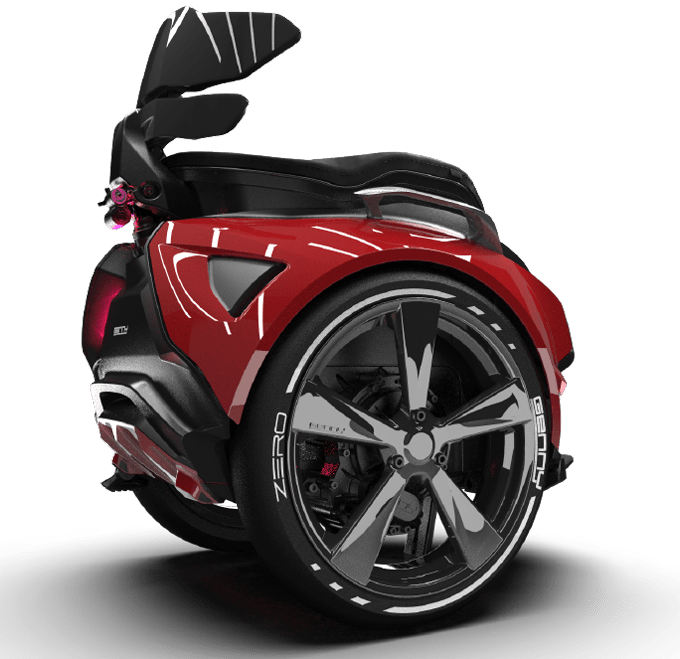}
    \includegraphics[width=0.495\linewidth]{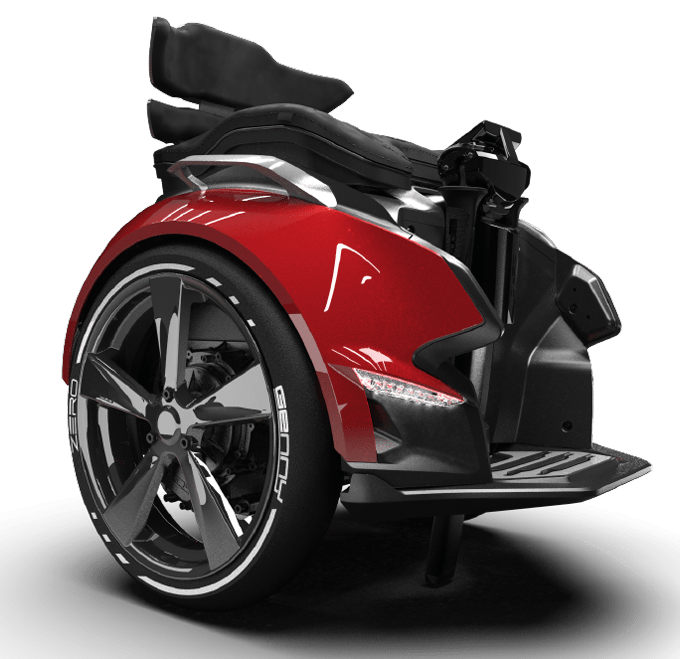}
    \caption{Front and back view of the original Genny Zero platform.}
    \label{fig:genny}
\end{figure}
Genny Zero\footnote{\url{https://www.gennymobility.com}} is a self-balancing powered wheelchair designed as a medical mobility device for individuals with disabilities, see Fig.~\ref{fig:genny}. Unlike conventional wheelchairs, which typically require manual propulsion or joystick-based control, Genny Zero is operated mainly through body-weight shifting. This allows the user to control the longitudinal motion of the platform in an intuitive and hands-free manner.
% As a result, the users gain great independence, as their hand are freed for other activities, thereby enhancing their quality of life.

%
The operating principle of Genny Zero is analogous to that of other self-balancing systems~\cite{Brentari:icm:2015} like Segway. 
The platform can be modeled as an inverted-pendulum system, where the main body, including the user seat, is mounted to the wheel axis. The body is free to rotate around this axis, while the wheels are actively controlled to maintain the upright equilibrium of the system. 
When the user (sitting on the main body) leans forward or backward, the equilibrium is perturbed. The controller reacts by driving the wheels so as to restore balance, and this corrective action generates the longitudinal motion of the wheelchair. 
In addition, Genny Zero is a non-holonomic platform: its planar motion is constrained by its wheel configuration, and it cannot move laterally. 
Steering is therefore achieved by controlling the yaw motion of the platform through the handlebar: the user moves the handlebar in the desired turning direction, while the handlebar angle determines the magnitude of the steering command. 
The combination of body-weight shifting for forward and backward motion and handlebar-based steering provides a compact and intuitive control interface.

The platform supports three operating modes. In \emph{driving} mode, the user sits on Genny Zero and has full control of the device, using body motion to regulate forward and backward motion, and the handlebar to steer. In \emph{leash} mode, a person can control the platform while not being seated on it. In this mode, Genny Zero remains self-balanced while standing still, and the person can move it by perturbing its equilibrium and steering through the handlebar. Finally, in \emph{software-control} mode, desired pitch and steering commands can be sent through software, enabling autonomous operation without direct user input. 
%
% In our implementation, the autonomy layer enforces an additional command-timeout safety mechanism: if no command is generated for $0.25$~s, the system sends a stop command and the platform continues self-balancing in place.
%

%
Genny Zero weighs $70$~kg and supports a payload of up to $120$~kg. It includes redundant safety components, such as dual-winding motors, dual batteries, and dual control boards, which support continued operation in case of component failure. 
The device can handle slopes of up to $20^\circ$, overcome small steps, and traverse different terrains, including sand and snow. Its maximum operating range is up to $30$~km, and its maximum velocity is up to $20$~km/h, corresponding to approximately $5.55$~m/s. 
For safety reasons, the specific platform used in this work is hardware limited to $10$~km/h ($\approx2.8$~m/s). In all experiments presented in this work the maximum velocity allowed by software was $1.0$~m/s. 
For the remainder of the paper, we refer to two main reference frames associated with the platform. 
The \texttt{base} frame is located at the midpoint of the wheel axis and is rigidly attached to the Genny Zero body; therefore, it pitches together with the platform. 
The \texttt{base\_footprint} frame is defined as the projection of the \texttt{base} frame onto the ground plane. As a result, \texttt{base\_footprint} follows the planar position and yaw of the platform, but does not include its pitch motion.
The inertial frame, useful to describe the motion of the device in the environment, is instead called \texttt{map} frame.

%
% For these reasons, Genny Zero provides an attractive platform for assistive autonomy, while also introducing specific integration challenges due to its self-balancing dynamics, non-holonomic motion constraints, and compact mechanical design.
%
\subsection{Sensory equipment}

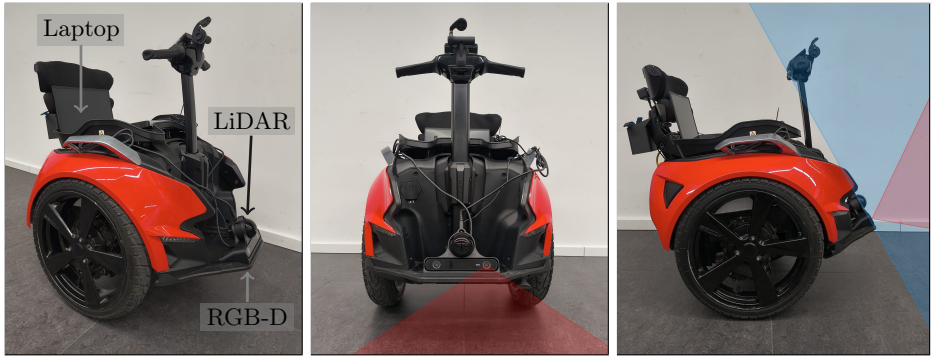
\begin{figure}[t]
    \centering
    \frame{\begin{tikzpicture}%
        \node[anchor=south west,inner sep=0] at (0,0){\includegraphics[height=0.38\linewidth, trim={0 0 0 6cm}, clip]{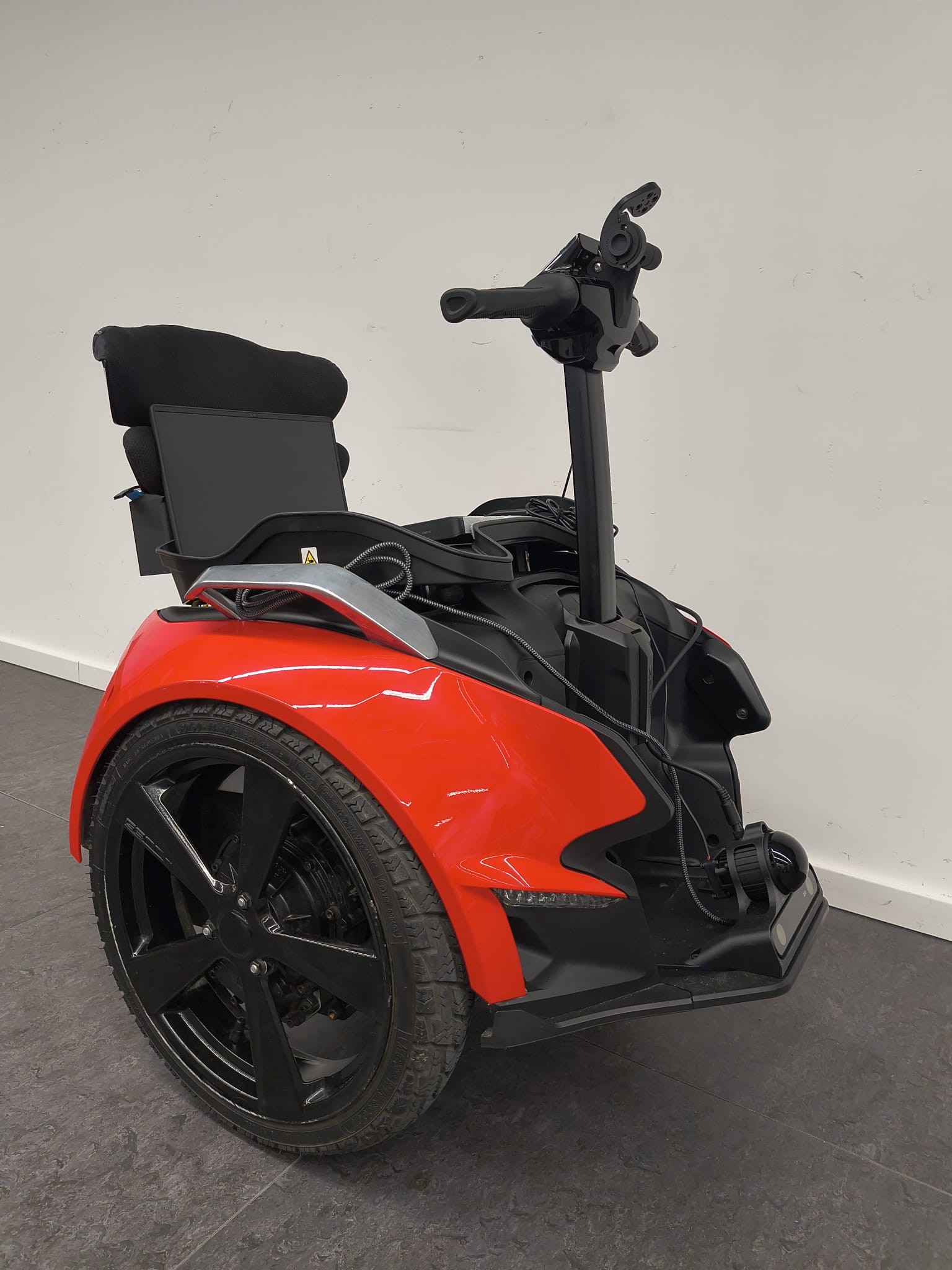}};%
        \draw[->,line width=1.0,color=black!50] (1.0,4.1) -- (1.0,3.2);
        \node[fill=black!50!white,
    fill opacity=0.3, text opacity=1, inner sep=2pt, text width=1.0cm, align=center,font=\footnotesize\selectfont] at (1.0,4.3) {Laptop};%
        \draw[->,line width=1.0,color=black!90] (3.25,2.9) -- (3.2,1.85);
        \node[fill=black!50!white,
    fill opacity=0.3, text opacity=1, inner sep=2pt, text width=1.0cm, align=center,font=\footnotesize\selectfont] at (3.25,3.1) {LiDAR};%
        \draw[->,line width=1.0,color=black!50!white] (3.2,0.7) -- (3.2,1.10);
        \node[fill=black!50!white, fill opacity=0.5, text opacity=1, inner sep=2pt, text width=1.1cm, align=center,font=\footnotesize\selectfont] at (3.2,0.5) {RGB-D};%
    \end{tikzpicture}}
    \hfill
    \frame{\includegraphics[height=0.38\linewidth, trim={0 6cm 0 0}, clip]{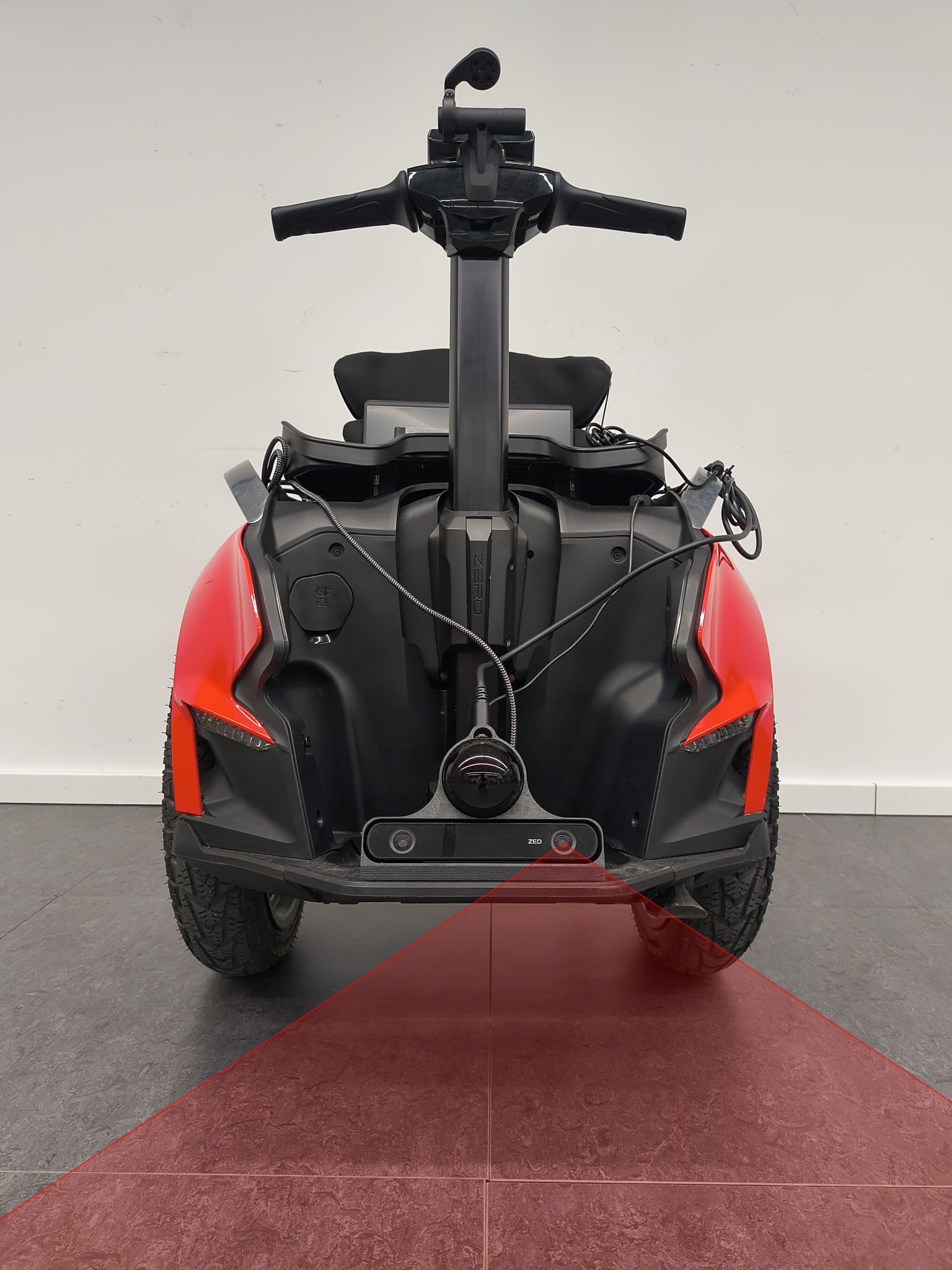}}
    \hfill
    \frame{\includegraphics[height=0.38\linewidth, trim={3cm 4cm 2cm 10cm}, clip]{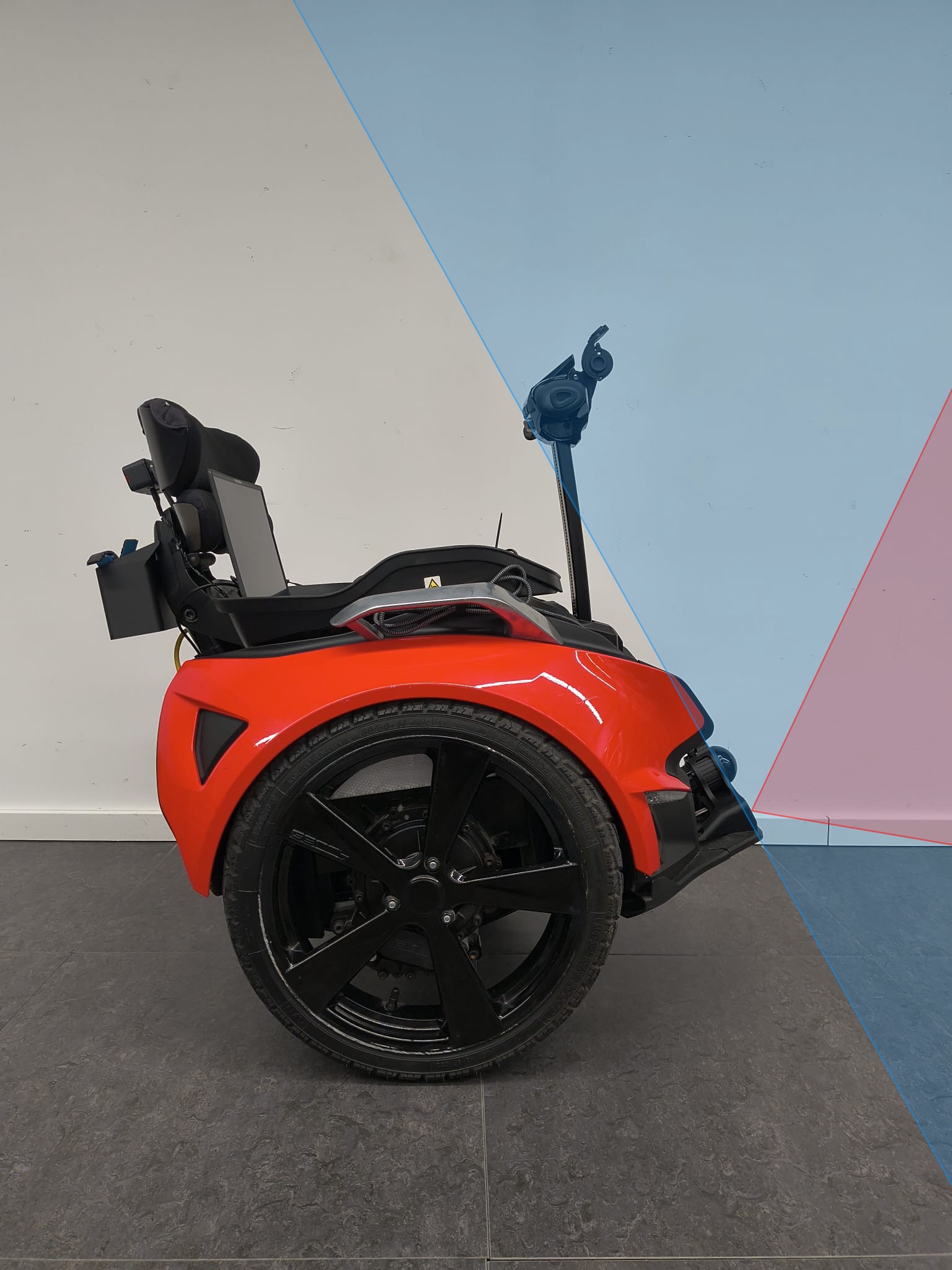}}
    \caption{3D, front and side views of the modified Genny Zero platform equipped with the LiDAR and RGB-D sensors, along with the laptop used for sensor acquisition and the navigation controller. The sensors' FOV are highlighted in red for the ZED~2i and blue for the Robosense Airy LiDAR.}
    \label{fig:genny_sensor_setup}
\end{figure}
To extend Genny Zero with perception capabilities, we equipped the platform with a frontal sensing module composed of two sensors: a Robosense Airy LiDAR and a ZED~2i stereo camera with a $2.1$~mm focal length. 
%
%This setup represents the current frontal-only sensing configuration of the platform.
%
The Robosense Airy is a hemispherical LiDAR with a field of view of $360^\circ$ horizontally and $90^\circ$ vertically in the sensor frame. 
The ZED~2i is used at HD720 resolution, for which it provides a field of view of $101^\circ$ horizontally and $68^\circ$ vertically. 
\rev{Both sensors operated at a frequency of $10$~Hz.}
The different views in Fig.~\ref{fig:genny_sensor_setup} illustrate the current sensor integration and highlight the corresponding fields of view and coverage around Genny Zero.
Both sensors and the Genny Zero platform are connected to an onboard laptop, which performs all computation required by the perception, navigation, and control modules. In the current experimental setup, the laptop is positioned in place of the user seat.
\rev{The onboard computer ran Ubuntu~22.04 and ROS~2 Humble, and is equipped with an AMD Ryzen~9~8945HS CPU, an NVIDIA GeForce RTX~4070 Laptop GPU, and $32$~GB of RAM.}
The two sensors play complementary roles in the current system. The LiDAR is primarily intended for localization and navigation tasks, including obstacle avoidance, estimation of ground-surface irregularities, and broad structural awareness of the surrounding environment.
The ZED~2i is mainly intended for tracking people in front of the wheelchair and for providing visual information useful for HRI. 
More generally, the goal of this design is to combine two sensing modalities with different strengths: the LiDAR provides wide-angle spatial awareness that is suitable for navigation, while the ZED~2i provides richer visual information that is more informative for interaction-related decisions.
The sensor mount is attached to the Genny Zero footrest using existing mounting holes. This choice was made to preserve the original and already validated structure of the platform, while integrating the additional sensing hardware with minimal mechanical changes. 
As a result of the low mounting position, the current sensor support is inclined by $20^\circ$ with respect to the footrest plane.  
With this configuration, the ZED~2i points upward, and its field of view intersects the ground plane at approximately $3.5$~m in front of the platform, while still allowing people tracking down to about $1$~m from the Genny Zero body frame. 
Similarly, the LiDAR is mounted with an additional $90^\circ$ rotation with respect to the ZED~2i, causing its field of view to intersect the ground plane at approximately $0.6$~m from the Genny Zero \texttt{base\_footprint} frame.

\subsection{Perception and navigation modules}

The autonomy stack is built around a custom Genny Zero driver package and the ROS~2 Nav2 navigation framework\footnote{\url{https://docs.nav2.org/}}. 
The Genny Zero driver uses a coarse kinematic model of the platform and interfaces with the internal control boards of the device. It reads the main platform states, including wheel velocities, pitch, and body velocity, and exposes the standard interfaces required by a robotic navigation pipeline, such as odometry and the TF tree. 
In the opposite direction, the driver receives velocity commands in the form of ROS Twist messages and converts them into the desired pitch and steering commands accepted by the Genny Zero platform. 

Longitudinal velocity control is implemented as an outer PID loop. This loop converts the desired linear velocity into a desired pitch command, while the internal Genny Zero control boards close the low-level balancing loop. 
The steering component of the incoming command is instead converted into the corresponding steering input for the platform. 
This separation allows the autonomy stack to command the platform using standard mobile-robot velocity interfaces, while preserving the internal self-balancing control structure of Genny Zero.

Mapping, localization, and navigation are implemented using Nav2. 
Since the Nav2 pipeline commonly operates on planar laser scans, the 3D LiDAR point cloud is converted into a virtual 2D scan. To this end, a dedicated node extracts a slice of points within a band parallel to the ground plane. The selected points are then resampled into angular bins of $0.5^\circ$, producing a virtual planar scan centered at the 3D LiDAR origin. In the default configuration, the scan covers the forward-facing semicircle from $-90^\circ$ to $90^\circ$. 
The virtual scan frame is kept aligned with gravity rather than with the pitching body of the platform. As a result, Nav2 receives a stable planar scan even though the Genny Zero body and the sensors mounted on it move with the self-balancing dynamics.

In the experiments presented in this paper, mapping is performed in environments without people. The resulting map is then saved and reused during execution. 
During autonomous operation, the system localizes against this known map and treats all additional perceived structures, including people, as temporary obstacles. 
The navigation pipeline uses Simultaneous Localization And Mapping (SLAM) for map construction, Adaptive Monte Carlo Localization (AMCL) for localization against the saved map, $A^*$ for global planning, and the Regulated Pure Pursuit Controller for path tracking~\cite{macenski2023regulated}. 
The velocity commands produced by Nav2 are sent to the Genny Zero driver as Twist commands and are then converted into platform-level pitch and steering commands.
\rev{At the Nav2 level, the platform is represented through the standard planar mobile-robot abstraction. AMCL is configured with a differential-drive motion model, and planning and control are performed in the \texttt{base\_footprint} frame, which removes pitch from the navigation representation. Therefore, although pitch is measured and exposed by the driver, it is not explicitly used by Nav2 for localization, planning, or local control. This approximation was sufficient for the low-speed proof-of-concept demonstrations presented in this work, but it does not explicitly model the self-balancing dynamics of the platform.} 
\rev{To keep the demonstrations conservative, the Nav2 configuration imposed velocity limits well below the nominal platform maximum speed of $20$~km/h. The Regulated Pure Pursuit controller was configured with a desired linear velocity of $0.65$~m/s, while the velocity smoother limited the command to $0.75$~m/s and $0.95$~rad/s, with linear acceleration and deceleration limits of $1.20$~m/s$^2$ and $-0.70$~m/s$^2$, respectively. The controller server was configured at $20$~Hz, the local costmap was updated at $5$~Hz, and the global planner was expected at $1$~Hz. The local and global costmaps used obstacle and inflation layers, with an inflation radius of $0.7$~m. Collision checking was enabled in the Regulated Pure Pursuit controller, and reverse motion was disabled. These settings supported cautious operation during the presented demonstrations; however, they do not constitute a complete safety validation and do not provide formal safety guarantees.}

\rev{In our implementation, the autonomy layer includes a command-timeout safety mechanism: if no command is generated for $0.25$~s, the system sends a stop command and the platform remains self-balanced in place. During the experiments, the system was always supervised by an operator, who could interrupt the autonomous behavior at any point from the operator computer by triggering this timeout mechanism. In the current prototype, the low-level internal Genny Zero balancing controller is not directly accessible, and the autonomy layer cannot send torque commands. Instead, it sends only velocity commands and relies on the internal controller to maintain self-stabilization. Deeper hardware integration with the Genny Zero platform is planned for future work, with the aim of enabling more advanced control schemes.}

\begin{figure}[t]
\centering
\frame{\includegraphics[width=0.45\linewidth, trim={0cm 2cm 20cm 1cm}, clip]{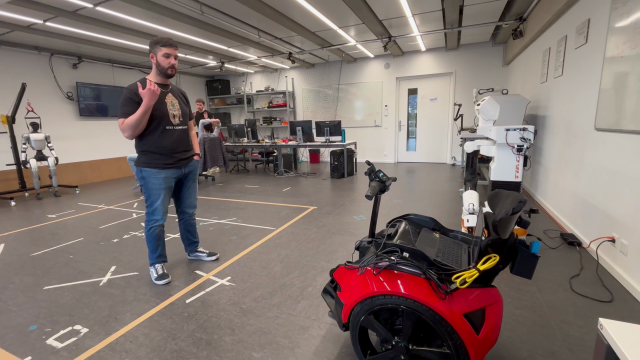}}
\quad
\frame{\includegraphics[width=0.45\linewidth, trim={2cm 2cm 18cm 1cm}, clip]{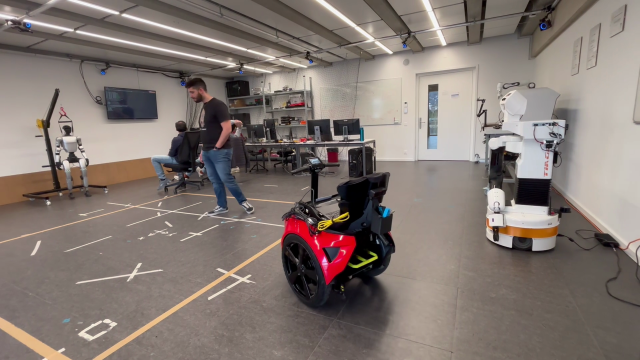}}
\\[10pt]
\frame{\includegraphics[width=0.45\linewidth, trim={2cm 2cm 18cm 1cm}, clip]{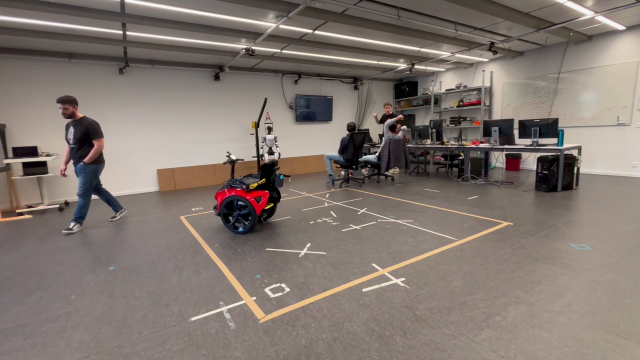}}
\quad
\frame{\includegraphics[width=0.45\linewidth, trim={2cm 3cm 18cm 0}, clip]{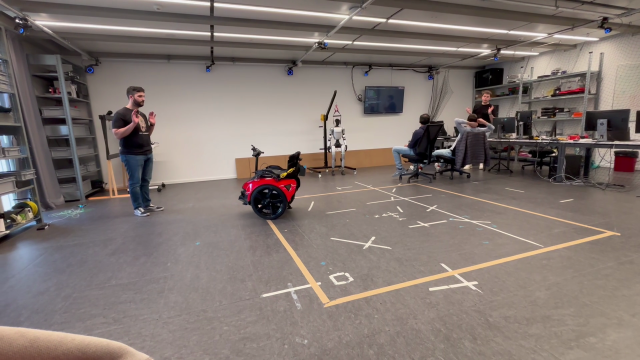}}
\caption{Representative snapshots of the gesture-based people-following behavior. The top-left image shows the \emph{follow me} gesture used to trigger people following, while the bottom-right image shows the \emph{stop} gesture used to stop the Genny Zero platform.}
\label{fig:people_following}
\end{figure}

The ZED~2i is currently used only for its 3D body tracking functionality\footnote{\url{https://www.stereolabs.com/docs/body-tracking}}. In particular, we use the accurate body tracking model, which provides 3D position estimates for $34$ body joints together with head orientation information.
The estimated body skeleton is processed by a gesture recognition module based on simple hands position rules. 
At the current stage, the system recognizes three gestures that can be used to trigger high-level behaviors. 
The first is \emph{hailing}, in which the user raises one hand above the head. 
The second is \emph{follow me}, in which the user brings one hand close to the corresponding shoulder. 
The third is \emph{stop}, in which the user places both hands close to the corresponding shoulders.
Examples of these gestures are shown in Fig.~\ref{fig:people_following} and Fig.~\ref{fig:hailing}, where they have been used to realized the real-world applications detailed in Sec.~\ref{sec:applications}.

% Overall, mapping, localization, and waypoint navigation are currently the most mature parts of the stack, while gesture-based interaction and platform-specific local planning remain under active development.
%
\section{Real-world applications}\label{sec:applications}

We evaluate the proposed system in two practical scenarios that are relevant for powered wheelchair assistance.
The first scenario is people-following. In this mode, the wheelchair follows a person, allowing the system to reproduce the trajectory selected by a human leader. This functionality is particularly relevant in structured environments, such as hotels and airports, where following a person or another vehicle may support efficient navigation. It is also useful in crowded or densely populated spaces, where tracking a skilled human leader can help the wheelchair follow socially appropriate paths.
The second scenario is hailing, in which the user summons the wheelchair from a distance. This behavior illustrates how autonomous functionality can extend the practical capabilities of a powered wheelchair, increasing convenience and supporting more flexible use of the device.
The experiments were conducted in our indoor laboratory and in the adjacent corridors of our university building. They are intended as preliminary demonstrations of the proposed autonomous functionalities \rev{rather than as a quantitative benchmark. We use the experiments to verify that the integrated perception, gesture-recognition, and navigation pipeline can execute the target behaviors on the real Genny Zero platform and to identify the main limitations that arise from this integration.}
In this setting, LiDAR data are fundamental for building the map of the environment, localizing the platform, and detecting temporary obstacles, thereby supporting safe navigation in populated environments allowing  the system to maintain collision-free motion in the mapped environment.
The accompanying video showing the complete execution of the presented experiments is available at: \url{https://youtu.be/LVAix_Qx7bM}.

\subsection{People-following}

In the first experiment, we demonstrate gesture initiated people following. The overall behavior is shown in Fig.~\ref{fig:people_following}.
The interaction starts when the user performs the \emph{follow me} gesture. This gesture is detected by the RGB-D-based perception module and interpreted as a command to activate the people following behavior.
After activation, the wheelchair locks onto the corresponding user and uses the available body tracking information to follow their motion. More specifically, the 3D position of the user's pelvis is projected onto the ground plane and transformed into the wheelchair \texttt{base\_footprint} frame. The resulting 2D position is used to define the line from the wheelchair to the user.
The navigation goal is selected on this line, offset from the user by the desired following distance of $1.25$~m.  Its orientation is aligned with the user-bearing direction, up to a small heading deadband. 
The goal is then transformed into the inertial \texttt{map} frame and sent to Nav2 as a goal pose.
The behavior is terminated when the active user performs the \emph{stop} gesture, detected as both hands being close to their respective shoulders for the configured hold duration.
\begin{figure}[t]
\centering
\frame{\includegraphics[height=0.35\linewidth, trim={18cm 2cm 18cm 5cm}, clip]{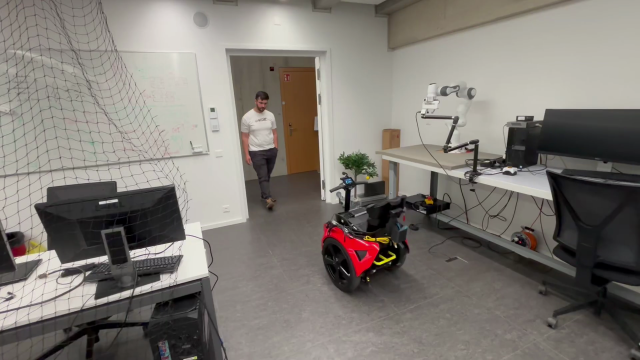}}
\hfill
\frame{\includegraphics[height=0.35\linewidth, trim={11cm 0 20cm 0}, clip]{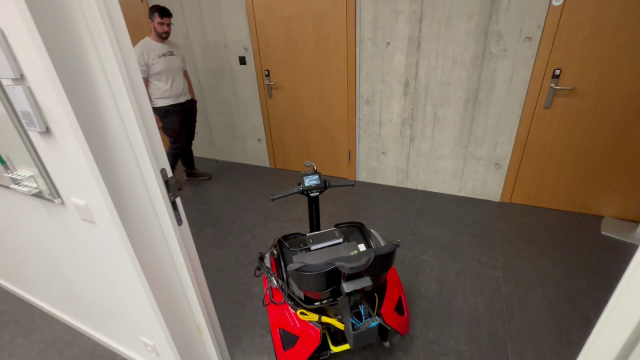}}
\hfill
\frame{\includegraphics[height=0.35\linewidth, trim={11cm 0 20cm 0}, clip]{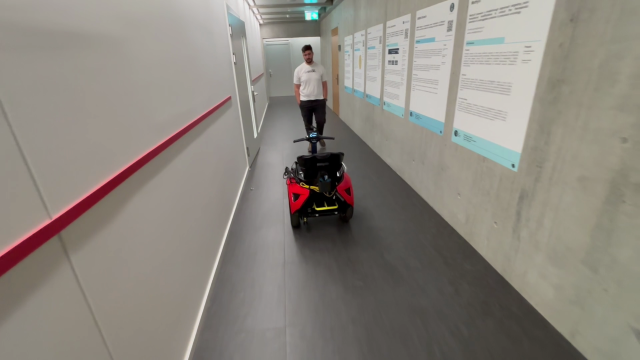}}
\hfill
\frame{\includegraphics[height=0.35\linewidth, trim={20cm 0 11cm 0}, clip]{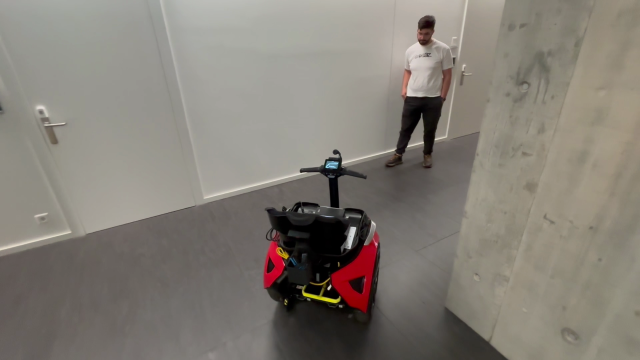}}
%\frame{\includegraphics[height=0.26\linewidth, trim={20cm 0 20cm 0}, clip]{figures/experiments/people_following_corridor_5.png}}
\caption{Representative snapshots of the people-following behavior in a constrained environment, including door crossing and corridor navigation.}
\label{fig:people_following_constrained}
\end{figure}
\begin{figure}[t!]
    \centering
    \includegraphics[width=0.7\linewidth]{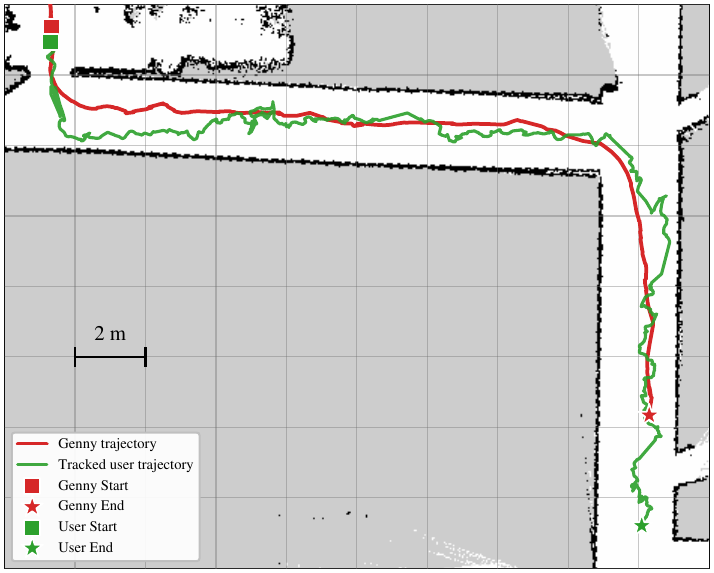}
    \caption{Top-down view of the trajectories recorded during the constrained indoor navigation experiment. The red trajectory represents Genny Zero, while the green trajectory represents the tracked user.}
    \label{fig:people_following_plot}
\end{figure}

In the second experiment, we extend the people-following behavior with obstacle avoidance, demonstrating its applicability to structured indoor navigation.
Figure~\ref{fig:people_following_constrained} shows four representative snapshots of the experiment.
The wheelchair follows the user while passing through a doorway, navigating along a corridor, and negotiating turns.
The corresponding top-down trajectories of Genny Zero and the followed user are shown in Fig.~\ref{fig:people_following_plot}.

\begin{figure}[t]
\centering
\frame{\includegraphics[height=0.20\linewidth, trim={0 0 0 0}, clip]{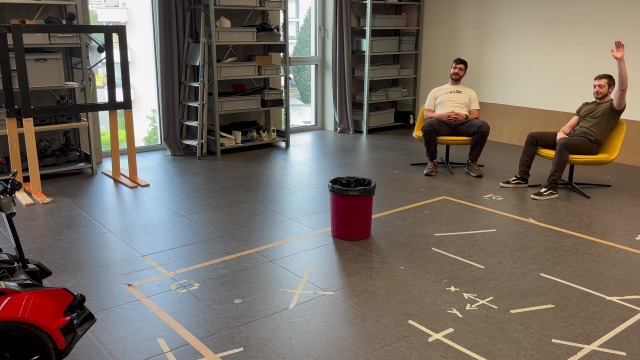}}%
\hfill
\frame{\includegraphics[height=0.20\linewidth, trim={0 2cm 0 2cm}, clip]{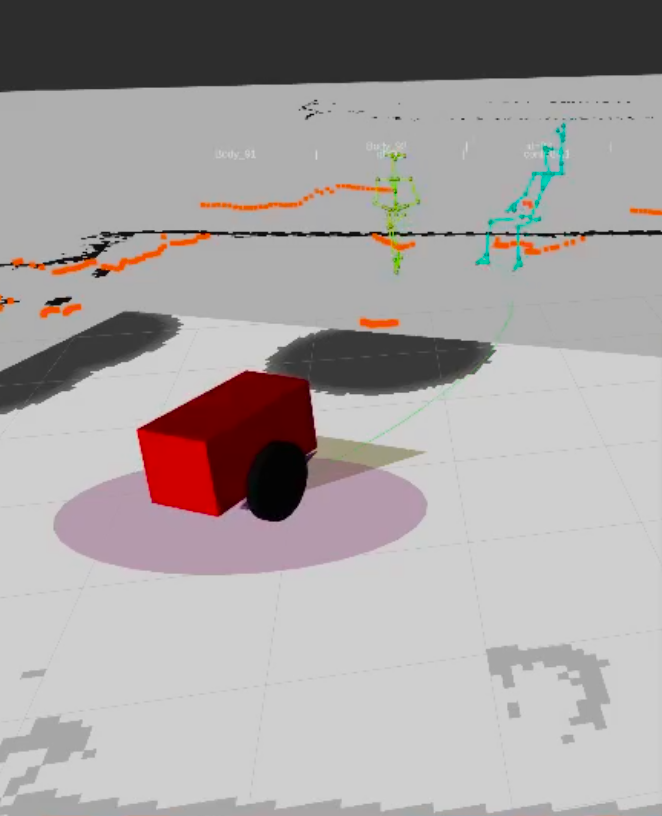}}%
\hfill
\frame{\includegraphics[height=0.20\linewidth, trim={0 0 0 0}, clip]{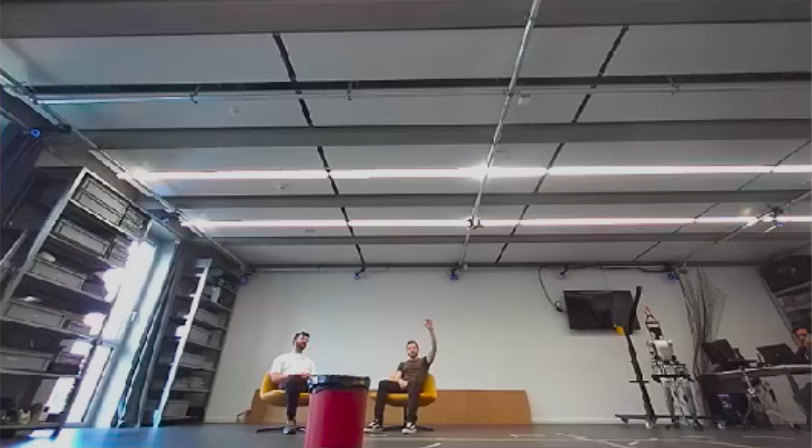}}%
\\[8pt]
\frame{\includegraphics[height=0.20\linewidth, trim={0 0 0 0}, clip]{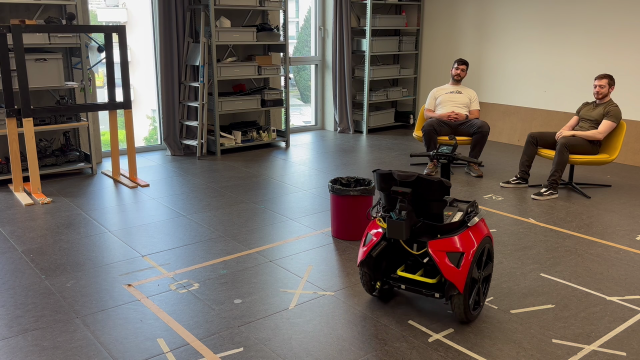}}%
\hfill
\frame{\includegraphics[height=0.20\linewidth, trim={0 2cm 0 2cm}, clip]{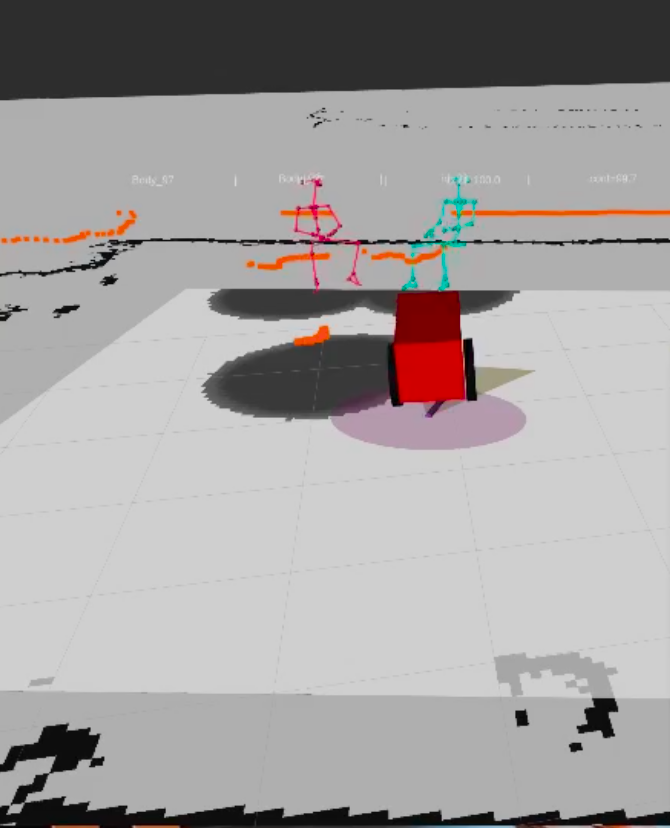}}%
\hfill
\frame{\includegraphics[height=0.20\linewidth, trim={0 0 0.15cm 0}, clip]{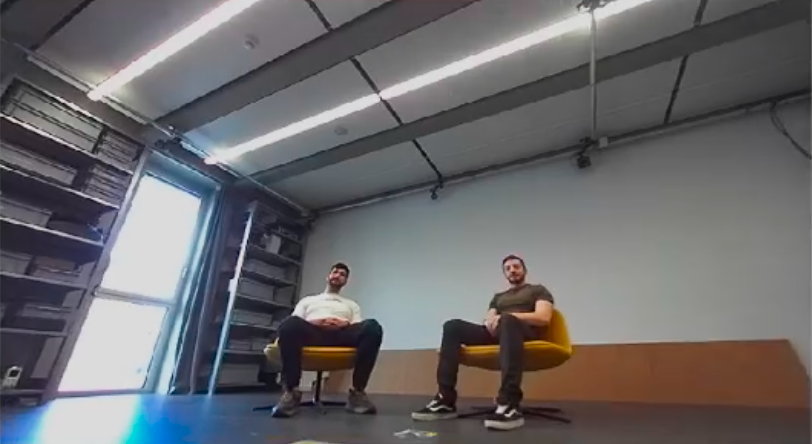}}%
\\[8pt]
\frame{\includegraphics[height=0.20\linewidth, trim={0 0 0.1cm 0}, clip]{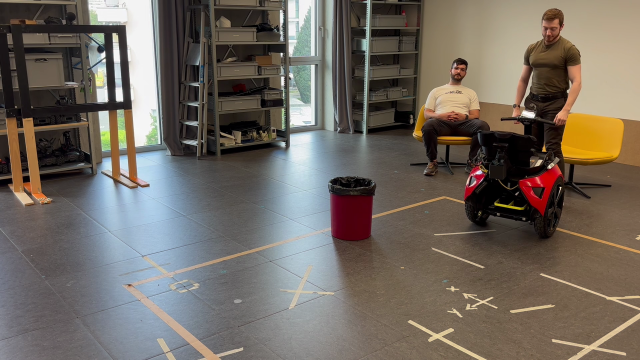}}%
\hfill
\frame{\includegraphics[height=0.20\linewidth, trim={0 2cm 0.15cm 2cm}, clip]{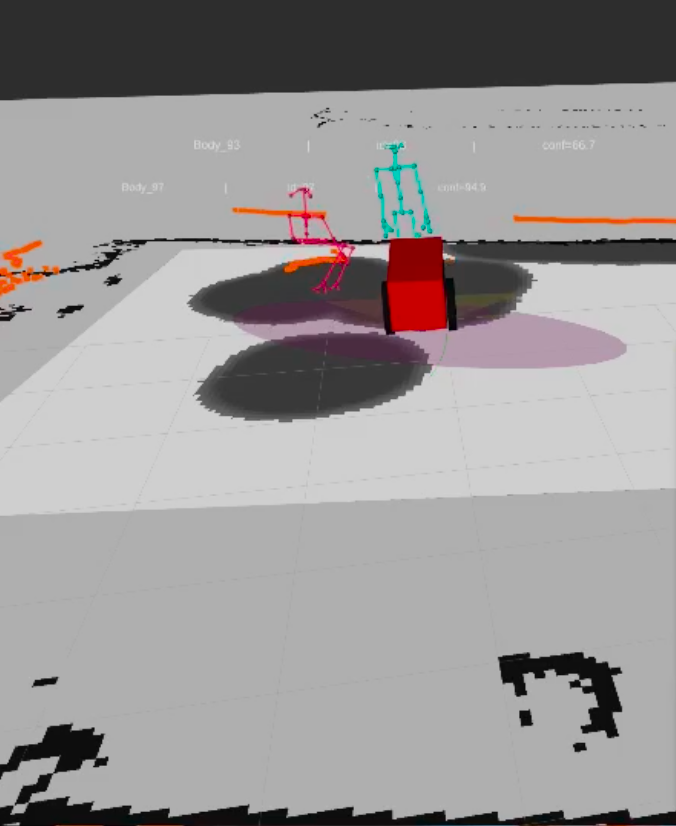}}%
\hfill
\frame{\includegraphics[height=0.20\linewidth, trim={0 0 0 0}, clip]{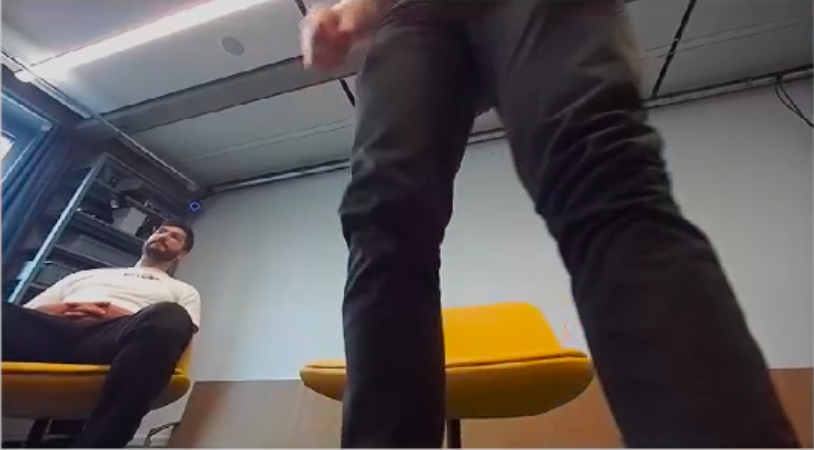}}
\caption{Representative moment of the hailing behavior (left), along with the corresponding LiDAR-based map used for obstacle avoidance (center) and the onboard RGB-D camera view (right). The top left frame shows an example of the \emph{hailing} gesture used to start the hailing behavior.}
\label{fig:hailing}
\end{figure}
\subsection{Hailing}
In the third experiment, we demonstrate hailing as a practical autonomous function that allows the user to call the wheelchair from a distance.
The behavior is triggered when the user performs the \emph{hailing} gesture by raising one hand above the head.
Once the gesture is detected, the RGB-D perception module localizes the user and select the target pose using the same procedure adopted for people-following, with a desired final distance of $1.0$~m from the user.
\rev{In the demonstrated run, the hailing gesture was performed at approximately $5$~m from the wheelchair.}
The navigation stack then generates a collision-free trajectory toward this pose while accounting for obstacles along the path. 
Once at the correct distance to the user, the wheelchair finally stops and goes in \emph{leash} mode giving control to the user.
Figure~\ref{fig:hailing} illustrates three key moments of the experiment. Each row reports the external view of the behavior, the corresponding LiDAR-based map, and the onboard RGB-D camera view.

\section{Towards real-world assistive mobility}~\label{sec:discussion}

The presented experiments show the feasibility of augmenting the Genny Zero powered wheelchair with robotic perception and navigation skills.
In particular, people-following and hailing demonstrate how an assistive mobility device can exploit autonomous functions to support practical user-centered and human-friendly behaviors.
At the same time, the current system should be interpreted as a preliminary prototype rather than as a complete autonomous wheelchair.
Its main value lies in bridging a real commercial mobility platform with perception, interaction, and navigation modules, while exposing the integration challenges that must be addressed before full real-world in-the-wild deployment.
\rev{The current prototype was demonstrated with the wheelchair unoccupied. For safety reasons, the autonomous setup limited the motor torque to $1$~Nm, which was not sufficient to support a seated user.}

A limitation of the current prototype is sensing coverage.
The presented setup uses a frontal sensing configuration, which was selected as a practical first step to validate autonomous control on the Genny Zero platform.
With this configuration, the autonomous behaviors are restricted to forward motion and rotations, while backward autonomous motion is avoided.
Extending the system toward full $360^{\circ}$ perception will require additional sensing on the rear and lateral sides of the platform~\cite{Arreghini:arso:2026}.
However, achieving omnidirectional perception is not only a matter of adding a single additional LiDAR sensor.
Placing a sensor high above the platform could improve coverage, but would also introduce drawbacks in terms of size, usability, aesthetics, and mechanical integration.
A more suitable solution is likely to require a deeper integration of multiple compact sensors with complementary fields of view, so as to cover the area around the wheelchair while preserving the form factor and usability of the original device.

The current sensor placement also reflects a compromise between perception requirements and product-integration constraints.
The frontal mount was obtained after multiple design iterations and was selected to balance hardware fit, sensor field of view, and future user convenience.
Several constraints influenced this choice, including preserving user comfort, avoiding interference during boarding and dismounting, protecting the sensors from collisions, maintaining a compact shape, and limiting modifications to the validated mechanical structure of the platform.
This low frontal placement is useful for observing the region immediately in front of the wheelchair, including nearby pedestrians, legs, and low obstacles that are relevant for short-range collision avoidance.
At the same time, it introduces perception limitations, especially when people are close to the platform, such as occlusions reducing long-range visibility, sensitivity to slopes and curbs, and limited visibility of upper-body cues.

The current navigation pipeline relies on planar information reconstructed from LiDAR data.
\rev{This solution is effective for the preliminary demonstrations presented in this paper, but it has important validity limits. The navigation stack treats the wheelchair as a planar mobile base controlled through a quasi-static approximation, while the Genny Zero is a self-balancing platform whose longitudinal motion is coupled with body pitch. As a result, the current control interface is suitable for low-speed proof-of-concept operation, but it is not sufficient for user-ready autonomous mobility.}
\rev{A concrete example of this limitation emerged during people-following. Moderate noise and jitter in the estimated position of the tracked user caused frequent updates of the navigation goal. In some cases, these updates led the local planner to briefly stop while replanning. During these intervals, the Genny Zero received a zero-velocity command and was then commanded to move again immediately afterwards. This stop-and-go behavior introduced oscillations in the self-balancing platform, which could occasionally affect perception and tracking stability. These effects are less critical on statically stable mobile bases, where short command interruptions are naturally damped by the platform dynamics. On the Genny Zero, instead, transient command changes are more visible because they interact directly with the self-balancing dynamics.}
\rev{Future work should therefore explicitly account for the platform dynamics in the design of both the local planner and the control interface. In particular, we will investigate platform-aware controllers, such as LQR or MPC formulations, potentially combined with safety filters such as Control Barrier Functions. Such controllers should jointly consider pitch dynamics, command smoothness, balance preservation, and collision avoidance, rather than treating these aspects as separate modules.}
\rev{A further direction is the integration of richer 3D information for local perception and obstacle avoidance. Although the current prototype uses 3D LiDAR data to generate a virtual planar scan for Nav2, the full 3D structure of the environment is not exploited by the local controller. Integrating 3D LiDAR information within an obstacle-avoidance strategy tailored to the kinematics and self-balancing dynamics of the Genny Zero could improve navigation performance, especially in cluttered environments and in regions not covered by the front-facing RGB-D camera.}

Human-aware perception is indeed a central element of our roadmap.
The current RGB-D perception module enables gesture-based interaction and body tracking, which are sufficient for the demonstrated hailing and people-following behaviors.
However, more robust real-world operation will require handling more complex situations.
For instance, long-range user detection can be affected by occlusions, while final pose selection near the user must account for reachability, ergonomics, and accessibility for people with reduced mobility.
The same sensing system can also support richer HRI capabilities, such as reasoning about passerby intentions~\cite{Arreghini:iros:2024} and mutual gaze~\cite{Arreghini:hri:2024, Arreghini:icra:2024}.
These capabilities are all relevant for the challenging problem of social navigation in assistive mobility~\cite{Mavrogiannis:thri:2023}.

The experiments in this paper consider autonomous behaviors with the wheelchair unoccupied.
A fully assistive system must also operate when the user is onboard. In this case, autonomy should support the driver rather than replace them.
This motivates a shift from full autonomy toward shared-control and driver-assistance functions.
Examples include forward obstacle awareness, pedestrian-aware speed modulation, emergency stopping near hazards, corridor centering, doorway assistance, curb or drop-off awareness, and shared-control interventions in risky situations.
Such functions must be designed to preserve a satisfactory driving experience while improving safety and reducing the cognitive and physical burden on the user.

Finally, broader validation remains necessary.
The present results are based on early demonstrations in limited indoor environments and with partial sensing coverage.
Future work will require systematic evaluation across multiple users, environments, and interaction conditions.
This validation should assess not only navigation performance, but also usability, user acceptance, safety, and robustness.
Substantial engineering effort will also be required to translate the current prototype and future autonomous functionalities into a market-ready system suitable for everyday deployment.

\section{Conclusion}\label{sec:conclusion}
In this paper, we presented a preliminary study toward an autonomous powered wheelchair for real-world assistive mobility.
The proposed system is based on the Genny Zero, a commercially available self-balancing powered wheelchair that is not originally equipped with exteroceptive sensing or autonomous navigation capabilities.
We augmented the platform with RGB-D and LiDAR sensors and integrated the corresponding perception and navigation modules within a ROS~2-based architecture.

We demonstrated the prototype in two representative assistive scenarios: people-following and remote hailing.
These experiments show how robotic perception, human-aware interaction, and autonomous navigation can be combined on a real assistive mobility platform.
%
% The results support a broader roadmap toward intelligent driver-assistance functions for powered wheelchairs in everyday environments, including richer sensing coverage, more robust human-aware perception, shared-control strategies, and larger-scale validation with users.
\rev{The results support a broader roadmap toward intelligent driver-assistance functions for powered wheelchairs in everyday environments. The current system is a proof-of-concept prototype, but it provides useful guidance for future developments. In particular, future work must address platform-aware control, stronger safety mechanisms, omnidirectional sensing, dynamic-obstacle handling, shared-control strategies, and larger-scale validation with users before the system can be considered suitable for cluttered scenarios or crowded real-world operation.}
\begin{credits}
\subsubsection{\ackname} This work was supported by the Innobooster New Mobility program.
\end{credits}

\bibliographystyle{splncs04}
\bibliography{bibliography}

\end{document}